\documentclass[letterpaper, 10 pt, conference]{ieeeconf}  % Comment this line out if you need a4paper

\IEEEoverridecommandlockouts                              % This command is only needed if 
\usepackage{times}
\usepackage{epsfig}
\usepackage{graphicx}
\usepackage{amsmath}
\usepackage{amssymb}
\usepackage{xspace}
\usepackage{float}

\usepackage{booktabs}
\usepackage{mathtools}
\usepackage{xcolor}

\usepackage{amsfonts}

\usepackage{epstopdf}
\usepackage{color}

\usepackage{multirow}
\usepackage{pbox}

\usepackage{amsopn}
\usepackage{caption}
\usepackage{graphicx}
\usepackage{subcaption}
\usepackage{multirow}
\usepackage{longtable}
\usepackage{array}
\usepackage{makecell}

\usepackage{tikz}
\usepackage[table]{xcolor}
\usepackage{hyperref}

\DeclareRobustCommand\onedot{\futurelet\@let@token\@onedot}
\def\@onedot{\ifx\@let@token.\else.\null\fi\xspace}

\newcommand{\tableline}{\noalign{\hrule height 1.5pt}} % Thick

\definecolor{ubpubColor}{rgb}{0.43, 0.5, 0.5}

\newcommand{\subsubsubsection}[1]{
	\par\vspace{0.0cm}\textbf{#1:}\quad
	\ignorespaces
}
\newcommand{\CP}[1]{\ignorespaces}

\title{\LARGE \bf
Seg2Track++: Probabilistic Track Validation and  Data Association for Multi-Object Tracking and Segmentation
}

\author{D. Mendonça, T. Barros,  C. Premebida, U.J. Nunes % <-this % stops a space%
\thanks{Authors are with the University of Coimbra, Institute of Systems and Robotics, Department of Electrical and Computer Engineering, Portugal. Emails: \footnotesize\{diogo.mendonca,~tiagobarros,~urbano,~cpremebida\}@isr.uc.pt.}% <-this % stops a space
}

\begin{document}

\maketitle
\thispagestyle{empty}
\pagestyle{empty}

%%%%%%%%%%%%%%%%%%%%%%%%%%%%%%%%%%%%%%%%%%%%%%%%%%%%%%%%%%%%%%%%%%%%%%%%%%%%%%%%
\begin{abstract}
Autonomous systems require robust Multi-Object Tracking and Segmentation (MOTS) to operate reliably in dynamic environments, ensuring consistent object identities and precise mask-level delineation. Foundation models such as SAM2 have shown strong zero-shot generalization for segmentation, but their direct application to MOTS is limited by unreliable track association and false-positive propagation. This work introduces Seg2Track++, a framework that integrates instance segmentation with SAM2 and a novel track management module to perform zero-shot MOTS with enhanced temporal consistency. Tracks are associated using Mask Centroid Distance (MCD) and Confidence-Aware Cost Modulation (CCM), while Probabilistic Track Validation (PTV) employs a Bernoulli filter to validate track existence and suppress ghost tracks. Experimental results on KITTI MOTS demonstrate improved identity preservation, reduced false-positive propagation, and robust track management without fine-tuning.
\end{abstract}

%%%%%%%%%%%%%%%%%%%%%%%%%%%%%%%%%%%%%%%%%%%%%%%%%%%%%%%%%%%%%%%%%%%%%%%%%%%%%%%%
\section{INTRODUCTION}
Multi-Object Tracking and Segmentation (MOTS) is a critical perceptual capability for autonomous systems operating in complex, dynamic environments, requiring both the consistent preservation of object identities over time and accurate instance-level spatial segmentation \cite{voigtlaender2019mots}. Unlike conventional multi-object tracking (MOT), which primarily associates bounding-box trajectories, MOTS requires consistent instance masks under challenging conditions such as occlusions, appearance variation, motion blur, and dynamic backgrounds.

Modern MOT pipelines often follow the tracking-by-detection (TBD) paradigm, separating object detection from temporal association \cite{Sun2021}. While this modular design enables scalability, it also introduces vulnerabilities, including association errors and spurious track initialization, especially in crowded scenes or under prolonged occlusion \cite{Wang2025}.

Transformer-based foundation models for vision, such as Segment Anything Model 2 (SAM2) \cite{ravi2024sam2segmentimages}, have greatly advanced instance segmentation \cite{Xiangtai2024}. When integrated into MOTS pipelines \cite{jiang2025sam2mot}, SAM2’s rich appearance embeddings enhance identity preservation. However, erroneous segmentations may still be instantiated as valid tracks and persist across frames \cite{mendonça2025}, leading to false-positive propagation that degrades detection accuracy.

False-positive propagation remains a major source of failure in both MOT and MOTS systems \cite{Sadjadpour2024}. Spurious detections or segmentation masks can be temporally reinforced, persisting across frames and degrading association quality through identity switches, track fragmentation, and erroneous scene interpretation \cite{Gao2024}. These effects are especially problematic in safety-critical and long-term autonomous applications, highlighting the need for tracking approaches that better account for uncertainty and temporal consistency.

This work presents Seg2Track++, a zero-shot MOTS pipeline that addresses persistent challenges in segmentation-centric tracking, including false-positive propagation and association errors. The framework integrates enhanced data association strategies and a temporal probabilistic track validation mechanism, enabling more reliable track management and early suppression of spurious tracks. Evaluation on the KITTI Multi-Object Tracking and Segmentation (KITTI MOTS) benchmark demonstrates consistent improvements in tracking robustness and identity preservation over the original Seg2Track-SAM2.

In more details, the key contributions of this work are the following:

\begin{itemize}
    \item Enhanced data association for mask-based MOTS: Integration of Mask Centroid Distance (MCD) and Confidence-Aware Cost Modulation (CCM) strategies adapted for instance mask representations, improving correspondence accuracy under occlusion, motion ambiguity, and appearance variation.
    \item Probabilistic track validation (PTV): A Bernoulli filter-based mechanism that combines detector and SAM2 mask confidence to compute the posterior probability of track validity, enabling early detection and suppression of ghost tracks.
    \item Comprehensive evaluation on KITTI MOTS: Demonstration of improved tracking robustness, reduced false-positive propagation, and enhanced identity preservation, validating the effectiveness of the proposed temporal probabilistic and association mechanisms in a zero-shot SAM-based MOTS setting.
\end{itemize}

% % % ======================= New Section
\section{RELATED WORK}

False-positive propagation is a pervasive failure mode in multi-object tracking systems, arising from fundamental limitations in data association and track management that have been extensively studied in the tracking literature \cite{Sadjadpour2024, Zaech2022}. In crowded scenes or under partial and prolonged occlusions, spurious detections and erroneous associations can be temporally reinforced, allowing incorrect hypotheses to persist and contaminate subsequent matching decisions \cite{Wang2023}. This effect is amplified in MOTS, where high-fidelity instance masks may lend excessive confidence to false tracks once initialized \cite{mendonça2025}.

\begin{figure*}[t]
    \centering
    \includegraphics[width=\textwidth, trim={0cm 0cm 0cm 0.0cm},clip]{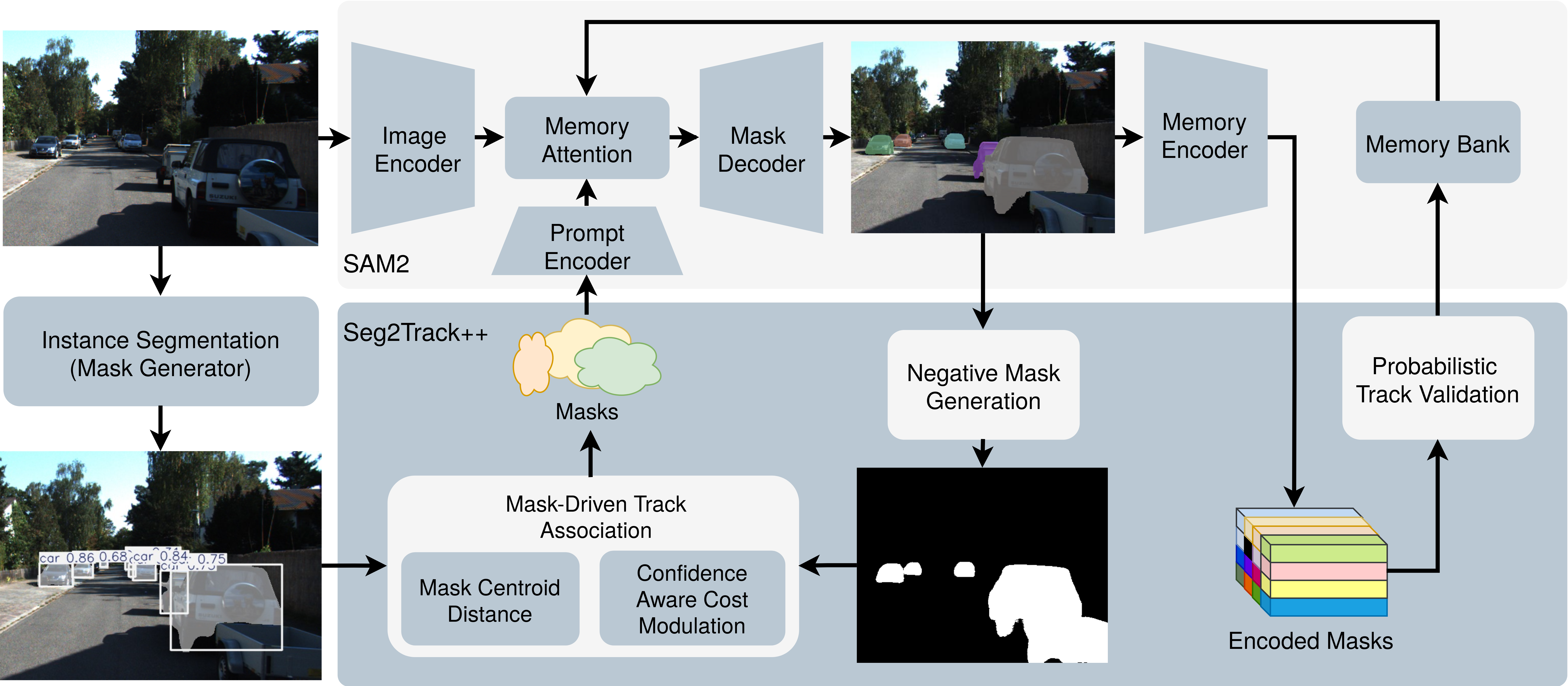}
    \caption{Illustration of the Seg2Track++, a framework for multi-object tracking and segmentation (MOTS) that integrates an instance segmentation model, SAM2, and a track management module. Seg2Track++ manages object tracks over time, initializing new tracks, reinforcing existing ones, and suppressing ghost tracks through mask-driven association and probabilistic temporal validation.}
    \label{fig:pipeline}
\end{figure*}

A central weakness identified in earlier association pipelines is the reliance on staged matching strategies that favor high-confidence detections, even when lower-confidence candidates exhibit better localization accuracy or temporal consistency \cite{Meng2025}. Recent works address this issue by adopting unified association schemes that evaluate all candidate assignments jointly, reducing early commitment errors and improving association stability \cite{Stadler2023}. These strategies are further strengthened by richer similarity formulations that more effectively integrate motion and appearance cues, including combined distance metrics and representations incorporating object shape, Mahalanobis distance, and soft bounding-box IoU \cite{stanojevic2024boosttrack++}. Together, these advances improve discrimination between true object hypotheses and spurious detections, forming the basis for more reliable tracking under occlusion and ambiguity \cite{Stadler2023,stanojevic2024boosttrack++}.

Beyond association accuracy, duplicate detections under occlusion can trigger erroneous track initialization, motivating occlusion-aware initialization mechanisms that suppress redundant track births \cite{Stadler2023}. In both 2D and 3D tracking scenarios, explicit track validity modeling and observational gating strategies have been introduced to reduce ghost trajectories, particularly for distant or partially occluded objects where detection confidence is unreliable \cite{Nagy2025RobMOT}. Related work explores principled Bayesian formulations of object birth, survival, and death, incorporating measurement-driven adaptive birth processes and detection probability modeling to improve robustness against spurious tracks \cite{ding2025optipmb}.

Further improvements are achieved through adaptive confidence handling and robust state estimation \cite{Wu2022, Liu2024}. Soft detection confidence boosting techniques allow low-confidence detections to contribute to tracking without reintroducing multi-stage association biases, while adaptive similarity thresholds improve robustness when tracklets are sparsely updated. In parallel, recent studies highlight the limitations of assuming precise detection localization in state estimation, noting that unmodeled measurement noise leads to state drift during prolonged occlusions \cite{Nagy2025RobMOT, Li2025}. Refined Kalman filtering formulations that explicitly account for localization uncertainty have been shown to improve state stability and recovery performance, contributing to consistent gains in MOTA, HOTA, and IDF1 across benchmark datasets \cite{Nagy2025RobMOT}.

% % % % % ================================ New Section
\section{METHODOLOGY}

The proposed framework addresses MOTS, defined as detecting, segmenting, and tracking multiple objects across a sequence of $T$ frames, ${I_1, I_2, \dots, I_T}$, where each frame $I_t \in \mathbb{R}^{H \times W \times C}$ has height $H$, width $W$, and $C$ channels. The objective is to predict $K$ object trajectories, $T = {T_1, \dots, T_K}$, with each trajectory $T_k = \{(t, m_{t,k}, s_{t,k}, c_{t,k},\text{id}_k)\}_{t \in F_k}$ containing a binary mask $m_{t,k} \in \{0,1\}^{H \times W}$, a confidence score $s_{t,k} \in [0,1]$, an occlusion score $o_{t,k} \in [0,1]$, a class label $c_{t,k}$, and a unique identity $\text{id}_k$. $F_k \subseteq {1,\dots,T}$ denotes the frames in which object $k$ is visible.

% Similarly to a previous work \cite{mendonça2025}, our framework, illustrated in \ref{fig:pipeline}, 
Seg2Track++, illustrated in \ref{fig:pipeline}, which leverages previous work \cite{mendonça2025}, integrates three components: (i) an object proposal model, in this case, an instance segmentation model, (ii) SAM2, responsible for transformer and mask-based tracking using attention mechanisms, and (iii) the Seg2Track++ module, which manages track initialization, reinforcement, and refinement.

\subsubsection{Object Proposal Model}
A state-of-the-art instance segmentation model produces mask proposals for all objects in each frame:
\begin{equation}
P_t = {p_{t,1}, \dots, p_{t,N_t}}, \quad p_{t,j} \in \mathbb{R}^2.
\end{equation}
These proposals are used to prompt SAM2 and are processed by Seg2Track++ for track management. Each new track inherits the class label of its initializing detection, as SAM2 is class-agnostic.

\subsubsection{Segment Anything Model 2}
SAM2 operates as a foundation model for mask-based tracking. Given frame $I_t$ and proposals $P_t$, it outputs
\begin{align*}
M_t = \Bigl\{ \bigl\{ (m_i, s_i, e_i,\text{id}_i) \bigr\}_{i=1}^{N_m} \mid & N_m \geq 0, \\
& m_i \in [0,1]^{H \times W}, \\
& s_i \in [0,1], \\
& o_i \in [0,1], \\
& e_i \in \mathbb{R}^{H_E \times W_E}, \\
& \text{id}_i \in  \{1, 2, . . . , K\} \Bigr\},
\end{align*}
where each tuple includes a soft segmentation mask $m_i$, IoU-based confidence score $s_i$~\cite{ravi2024sam},occlusion score $o_i$, encoded mask $e_i$ (downsampled via SAM2’s memory encoder), and identity $\text{id}_i$. 
Hence, for each frame $I_t$ and the corresponding bounding box prompts $P_t = \{p_{t,i}\}_{i=1}^{N_P} \in \mathbb{D}$ , SAM2 outputs a set of output vectors $M_t = \{(m_{t,i}, s_{t,i}, o_{t,i} e_{t,i}, \text{id}_{t,i})\}_{i=1}^{N_t} \in \mathbb{M}$.

3) Seg2Track++ extends the original Seg2Track module \cite{mendonça2025} while achieving the same goal: managing object tracks across frames through initialization, reinforcement, and refinement. Building upon the foundational capabilities of the previous work, the proposed framework introduces enhanced strategies for robust track association and probabilistic detection of false positives, which are detailed in the following sections.

\subsection{Mask-Driven Track Association and Initialization}

This module governs track initialization and association by determining which object proposals are used to prompt SAM2 at each frame. It operates in two sequential stages: (i) candidate selection for new track initialization via negative mask filtering, and (ii) association of remaining proposals with existing tracks using the proposed Mask Centroid Distance and Confidence-Aware Cost Modulation strategies.

\subsubsection{Track Initialization}

To identify proposals corresponding to previously unseen objects, a combined mask $N_{t-1}$ is constructed by aggregating all masks in $M_{t-1}$. Each proposal $p_{t,j}$ is evaluated based on its overlap with $N_{t-1}$:
\begin{equation}
O_j = \frac{|p_{t,j} \cap N_{t-1}|}{|p_{t,j}|}.
\end{equation}

Proposals with low overlap ($O_j < \tau_v$) and high segmentation confidence are treated as candidates for track initialization ($c_j > \tau_s$) and are forwarded to SAM2 to create new tracks. This step prevents re-initialization of already tracked objects while enabling the discovery of new instances. All remaining proposals are passed to the association stage.

\subsubsection{Track Association}
At each time step $t$, the goal of the association module is to assign instance segmentation proposals to existing object tracks in a temporally consistent manner. It is formulated as a bipartite matching problem between previous-frame track masks and current-frame proposals. For each possible pair $(m_{t-1,i}, p_{t,j})$, an association cost $C_{ij}$ is computed based on spatial and segmentation consistency. These costs form a matrix $C \in \mathbb{R}^{N_{t-1} \times N_d}.$ which is optimized globally using the Hungarian algorithm to obtain a one-to-one assignment.

Following assignment, each matched pair is validated using an overlap criterion. Specifically, a match is accepted only if the Intersection-over-Union (IoU) between the previous mask and the assigned proposal exceeds a minimum threshold $\tau_{\mathrm{IoU}}$. Pairs failing this validation are discarded, preventing spurious associations and limiting error propagation across frames. This general formulation provides a flexible framework in which different cost functions can be incorporated.

\subsubsubsection{Association with Mask Centroid Distance} While standard association strategies rely primarily on IoU-based costs, such formulations can be brittle under partial occlusions, mask fragmentation, or rapid object motion. To address these limitations, we propose a Mask Centroid Distance (MCD) cost that jointly models mask overlap and spatial consistency.

For each pair $(m_{t-1,i}, b_{t,j})$, the CD cost is defined as:
\begin{equation}
C_{ij} =
\alpha \left( 1 - \mathrm{IoU}(p_{t,j}, m_{t-1,i}) \right)
+
\beta \frac{\rho_{ij}^2}{c_{ij}^2},
\end{equation}
where $\rho_{ij}^2$ is the squared Euclidean distance between the centroids of the proposal and the previous mask, and $c_{ij}^2$ is the squared diagonal length of their smallest enclosing bounding box, serving as a normalization factor. The second term introduces a penalty for spatially distant candidates, ensuring that proposals far from the expected object location are discouraged even when partial overlap exists.

This formulation enables robust association in challenging scenarios where IoU alone is insufficient, such as during scale changes, fast motion, or temporary segmentation degradation.

\subsubsubsection{Confidence-Aware Cost Modulation} Beyond spatial and overlap consistency, segmentation confidence provides an important cue for association reliability. While low-confidence proposals can indicate false positives, discarding them can lead to track fragmentation, particularly under adverse conditions. To balance robustness and continuity, we introduce Confidence-Aware Cost Modulation (CCM).

Let $c_j \in [0,1]$ denote the segmentation confidence score of proposal $p_{t,j}$. If $s_j$ falls below a high-confidence threshold $\tau_s$, the corresponding association cost is scaled by a penalty factor $\gamma > 1$:

\begin{equation}
\tilde{C}_{ij} =
\begin{cases}
C_{ij}, & \text{if } c_j \geq \tau_c,\\
\gamma \cdot C_{ij}, & \text{if } c_j < \tau_c.
\end{cases}
\end{equation}

This mechanism does not exclude low-confidence proposals outright, but instead biases the assignment process toward higher-confidence matches when multiple candidates compete for the same track. As a result, CCM preserves the ability to recover from temporary segmentation uncertainty while substantially reducing the likelihood of erroneous associations driven by unreliable proposals.

\subsection{Probabilistic Track Validation}

To mitigate false-positive propagation, each track $i$ is assigned an existence probability $r_{t,i} \in [0,1]$, updated recursively via a Bernoulli filter incorporating detection confidence, SAM2 mask consistency, occlusion, and temporal decay. Matched tracks are reinforced based on evidence, attenuated by occlusion, while unmatched tracks decay over time. Tracks with $r_{t,i}$ below a threshold are discarded. This probabilistic mechanism provides principled temporal validation, suppressing ghost tracks while retaining uncertain but plausible tracks.

\paragraph{Bayesian derivation} The update function is derived directly from Bayes' theorem for a Bernoulli random variable. Let $H_i$ denote the hypothesis that object $i$ exists in the current frame, and $E_{t,i}$ the evidence provided by the frame, including detection confidence, mask overlap, and occlusion. Bayes' theorem gives the posterior probability:
\begin{equation}
r_{t,i} = P(H_i \mid E_{t,i}) = 
\frac{P(E_{t,i} \mid H_i) \, P(H_i)}{P(E_{t,i})}.
\end{equation}

Using the law of total probability, the marginal likelihood $P(E_{t,i})$ can be expanded as
\begin{equation}
P(E_{t,i}) = P(E_{t,i} \mid H_i) \, P(H_i) + P(E_{t,i} \mid \neg H_i) \, (1 - P(H_i)),
\end{equation}
where $P(\neg H_i) = 1 - P(H_i)$. Substituting, we obtain
\begin{equation}
\label{equ:bayes_post}
r_{t,i} = \frac{P(E_{t,i} \mid H_i) \, P(H_i)}{P(E_{t,i} \mid H_i) \, P(H_i) + P(E_{t,i} \mid \neg H_i) \, (1 - P(H_i))}.
\end{equation}

Defining the likelihood ratio of the evidence as
\begin{equation}
L_{t,i} = \frac{P(E_{t,i} \mid H_i)}{P(E_{t,i} \mid \neg H_i)},
\end{equation}
we can rewrite (\ref{equ:bayes_post}) in odds form as
\begin{equation}
\label{equ:bernoulli_update}
r_{t,i} = \frac{r_{t-1,i} \, L_{t,i}}{(1 - r_{t-1,i}) + r_{t-1,i} \, L_{t,i}},
\end{equation}
where $r_{t-1,i} = P(H_i)$ is the prior existence probability of track $i$ from the previous frame.  

\paragraph{Likelihood computation} The likelihood ratio $L_{t,i}$ encodes the evidence provided by the current frame, adjusted for occlusion and temporal decay:
\begin{equation}
L_{t,i} =
\begin{cases}
\exp\!\Big(\alpha \big( w_s \, s_{t,i} + w_\text{IoU} \, \text{IoU}_{t,i} \big)\Big), & \text{$T_i$ matched},\\[0.5em]
\exp(-\lambda \, \Delta t) \, (1 - o_{t,i}) + o_{t,i}, & \text{$T_i$ unmatched},
\end{cases}
\end{equation}
where $\alpha$ is a scaling factor controlling reinforcement strength, $w_s$ and $w_\text{IoU}$ are weights for detection confidence and mask agreement, $\lambda$ is the temporal decay rate, $\Delta t$ is the number of frames since the last match, and $o_{t,i}$ is the occlusion score.

Overall, this Bernoulli filtering mechanism enables robust temporal track validation by combining detection confidence, mask consistency, and occlusion awareness, effectively suppressing false positives while preserving uncertain but plausible tracks.

% % % % ====================== New Section
\section{EXPERIMENTAL EVALUATION}
This section presents the evaluation framework\CP{for our proposed approach}, detailing the dataset benchmarks, implementation details, empirical results and discussion.

\subsection{Benchmark}
The experimental evaluation of the proposed approach is conducted on the KITTI Multi-Object Tracking and Segmentation (MOTS) dataset~\cite{voigtlaender2019mots}, a widely used benchmark for segmentation-centric multi-object tracking. The dataset comprises 21 training sequences and 29 test sequences, with dense pixelwise segmentation labels for each object, enabling evaluation of detection, tracking, and segmentation performance simultaneously.

\subsubsection{Evaluation Metrics}
Evaluation follows the official KITTI MOTS protocol, with methods ranked primarily by the Higher Order Tracking Accuracy (HOTA) metric~\cite{Luiten2020IJCV}. HOTA jointly evaluates detection accuracy (DetA), association accuracy (AssA), and localization accuracy (LocA), providing a balanced measure of tracking performance. Unlike earlier metrics such as MOTA or MOTP, HOTA averages performance across varying detection thresholds, mitigating bias toward either precision or recall and offering a robust indicator of overall system quality.

\subsection{Implementation Details}\label{sec:implementation}
This section details the implementation procedures used to obtain the reported results. 
Seg2Track++ operates under a zero-shot paradigm, requiring no dataset-specific fine-tuning. For instance segmentation, we employ YOLO11-seg applied directly to the KITTI sequences using pre-trained weights. Mask generation is performed with SAM2.1-large, leveraging its pre-trained parameters to produce accurate and consistent instance masks.

\subsection{Results}
This section presents and discusses the empirical results of our proposed method. We evaluate its performance on KITTI MOTS benchmark. An ablation study is conducted to systematically assess the contribution of each module within our framework. Quantitative results for the KITTI MOTS benchmark are presented in Table~\ref{tab:mots} and the findings from the ablation study are detailed in Table~\ref{tab:ableton_mots}.

\begin{table*}[thp]
\centering
\caption{Results on the 2D KITTI MOTS benchmark (for \textit{car} and \textit{pedestrian} classes). The superscript (in blue) indicates the column-wise ranking of the methods.}
 {\renewcommand{\arraystretch}{1.5}% for the vertical padding
\begin{tabular}{l|cccc|cccc}
\tableline
 & \multicolumn{4}{c|}{Car} & \multicolumn{4}{c}{Pedestrian} \\
 Method & HOTA $\uparrow$ & DetA $\uparrow$ & AssA $\uparrow$&  LocA $\uparrow$& HOTA $\uparrow$ & DetA $\uparrow$ & AssA $\uparrow$ &  LocA $\uparrow$\\
\hline
 ViP-DeepLab~\cite{Qiao_2021_CVPR} & 76.38{\color{blue}$^1$} & 82.70{\color{blue}$^1$}  & 70.93  & 90.75{\color{blue}$^1$}  & 64.31{\color{blue}$^1$} & 70.69{\color{blue}$^1$} & 59.48 &  84.40{\color{blue}$^1$} \\
 EagerMOT~\cite{9562072} & 74.66{\color{blue}$^2$} & 76.11 & 73.75{\color{blue}$^3$} & 90.46{\color{blue}$^2$} & 57.65& 60.30 & 56.19 & 83.65 \\
OPITrack~\cite{gullapalli2021opitrack} & 73.04 & 79.44{\color{blue}$^2$}  & 67.97 & 88.57 & 60.38{\color{blue}$^2$} & 62.45 & 60.05{\color{blue}$^3$} & 83.55 \\
 ReMOTS~\cite{yang2020remots} & 71.61  & 78.32 & 65.98  &  89.33 & 58.81 & 67.96{\color{blue}$^2$} & 52.38 & 84.18{\color{blue}$^2$}  \\ 
 SearchTrack~\cite{tsai2022searchtrack} & 71.46  & 76.76 & 67.12 &  88.08 & 57.63  & 63.66{\color{blue}$^3$} & 53.12 & 80.89\\
 MOTSFusion~\cite{luiten19arxiv}  & 73.63 & 75.44 & 72.39 & 90.29{\color{blue}$^3$} & 54.04 & 60.83 & 49.45 & 83.71{\color{blue}$^3$}  \\
 PointTrack~\cite{xu2020Segment}  & 61.95 & 79.38{\color{blue}$^3$}  & 48.83 & 88.52 & 54.44 & 62.29 & 48.08 & 83.28\\
 TrackR-CNN~\cite{voigtlaender2019mots} & 56.63 & 69.90 & 46.53 &   86.60  & 41.93 & 53.75 & 33.84 & 78.03 \\  \hline
 \textit{(No fine-tuning)} &  &  & &    &  &  &  &  \\
 Baseline (Seg2Track-SAM2)~\cite{mendonça2025} & 74.13  & 71.03 & 78.15{\color{blue}$^1$} & 89.69  & 60.00 & 56.61 & 65.86{\color{blue}$^1$} & 80.40 \\ 
 Ours (Seg2Track++)  & 74.56{\color{blue}$^3$}  & 73.41{\color{blue}$^8$} & 76.50{\color{blue}$^2$} & 87.24{\color{blue}$^9$}  & 60.20{\color{blue}$^3$} & 57.40{\color{blue}$^8$} & 65.44{\color{blue}$^2$} & 79.63{\color{blue}$^8$} \\ 
 \tableline
\end{tabular}
}
\label{tab:mots}
\end{table*}

The evaluation of Seg2Track++ on the KITTI MOTS benchmark test set demonstrates competitive performance for both car and pedestrian classes, as seen in table \ref{tab:mots}. The framework achieves HOTA scores of 74.56\% for cars and 60.20\% for pedestrians, operating entirely in a zero-shot setting with no task-specific fine-tuning and relying exclusively on pretrained models.

Compared to the Seg2Track-SAM2 baseline, Seg2Track++ shows consistent improvements in both detection and tracking metrics. Detection accuracy (DetA) increases by approximately 2.4 percentage points (pp) for cars and 0.8 pp for pedestrians, largely due to the Probabilistic Track Validation (PTV) module, which suppresses false-positive propagation while carefully preserving plausible tracks. Association accuracy (AssA) is slightly lower for cars and similar for pedestrians, reflecting a trade-off introduced by PTV: some tracks removed to prevent false positives may later be reinitialized, temporarily affecting association.

Slight decreases in localization accuracy (LocA) are observed relative to the baseline , again reflecting the effect of PTV on temporarily removed tracks. Despite these minor reductions, the overall HOTA improves for cars and remains stable for pedestrians, indicating that the combination of enhanced association and probabilistic validation leads to a more reliable tracking pipeline with reduced false-positive propagation and maintained identity consistency.

Overall, Seg2Track++ confirms the benefits of combining advanced association strategies (MCD and CCM) with probabilistic temporal filtering (PTV), delivering improved zero-shot MOTS performance over the Seg2Track-SAM2 baseline by balancing detection reliability, identity preservation, and ghost track suppression.

\begin{table}[t]
\centering
\caption{Ablation study on the KITTI MOTS training set, for \textit{car} and \textit{pedestrian} classes.}
 {\renewcommand{\arraystretch}{1.5}% for the vertical padding
\begin{tabular}{ccccccc}
\hline
\rowcolor{white}
Class & MCA & CCM & PTV & HOTA & DetA & AssA \\
\hline
\textit{Car} 
            & & & & 80.454	& 79.039 & 82.354  \\
              & \checkmark & & & 80.470 & 79.105 & 82.417  \\
              &  & \checkmark & & 80.479	& 79.113 & 82.426   \\
              &  &  & \checkmark & 80.537	& 79.252 & 82.397   \\
              & \checkmark & \checkmark & \checkmark & 80.709	& 79.408 & 82.585   \\
\hline
\textit{Pedestrian}             
            & & & & 63.588	& 60.886 & 66.97  \\
              & \checkmark & & & 63.633 & 60.804 & 67.147   \\
              &  & \checkmark & & 63.79 & 61.032 & 67.266   \\
              &  &  & \checkmark & 64.194	& 61.542 & 67.515   \\
              & \checkmark & \checkmark & \checkmark & 64.226	& 61.628 & 67.525   \\
\hline
\end{tabular}
}
\label{tab:ableton_mots}
\end{table}

The ablation results in Table \ref{tab:ableton_mots} show that each module incrementally improves HOTA, DetA, and AssA for both car and pedestrian classes, with the full combination reaching 80.71\% HOTA for cars and 64.23\% for pedestrians. Gains are especially notable in AssA, reflecting stronger identity preservation and more reliable track associations. These findings confirm the complementary effect of Mask Centroid Distance (MCD), Confidence-Aware Cost Modulation (CCM), and Probabilistic Track Validation (PTV) in enhancing detection and tracking robustness. Differences between ablation gains and test-set performance can be attributed to threshold tuning: parameters such as IoU for association and PTV existence probability are optimized on the training set, whereas real test sequences contain edge cases where these thresholds are suboptimal. Despite this, the trends remain consistent, demonstrating the modules’ effectiveness under real-world conditions.

\section{CONCLUSION}

This paper presents Seg2Track++, a zero-shot MOTS framework that extends the capabilities of SAM2 by incorporating Mask Centroid Distance (MCD), Confidence-Aware Cost Modulation (CCM), and Probabilistic Track Validation (PTV). The framework demonstrates improvements in both association accuracy and false-positive suppression compared to its predecessor, Seg2Track-SAM2. Evaluation on the KITTI MOTS benchmark highlights three key advantages: (i) competitive performance in a fully zero-shot setting, requiring no task-specific fine-tuning; (ii) enhanced identity preservation and reliable track reinforcement under occlusion, motion ambiguity, and segmentation uncertainty; and (iii) principled probabilistic control of track persistence, reducing ghost tracks while maintaining plausible object continuity. Together, these contributions establish Seg2Track++ as a robust, detector-agnostic, and scalable solution for MOTS in dynamic, real-world environments.

\section*{ACKNOWLEDGMENT}
This work has been supported by the project PharmaRobot (ref. COMPETE2030-FEDER-01478600), and funded by Fundação para a Ciência e a Tecnologia (FCT), Portugal. This work was also supported by the ISR-UC FCT grant UID/00048/2025 (DOI: 10.54499/UIDB/00048/2025).

\bibliographystyle{IEEEtran}
\bibliography{abbrev_short,refs}

\end{document}